\documentclass{article}

\usepackage[final]{neurips_2022}

\usepackage[utf8]{inputenc} 
\usepackage[T1]{fontenc}    
\usepackage{hyperref}       
\usepackage{url}            
\usepackage{booktabs}       
\usepackage{amsfonts}       
\usepackage{nicefrac}       
\usepackage{microtype}      
\usepackage{xcolor}         

\usepackage{subcaption}
\usepackage{algpseudocode}
\usepackage{todonotes}
\usepackage{amsmath}
\usepackage{algorithm}
\usepackage{float}

\title{(When) Are Contrastive Explanations of Reinforcement Learning Helpful?}

\author{ \And
  Sanjana Narayanan \\
  Harvard University \\
  \texttt{sanjana\_narayanan@college.harvard.edu} 
  \And
  Isaac Lage \\
  Harvard University \\
  \texttt{isaaclage@g.harvard.edu }
  \And
  Finale Doshi-Velez \\
  Harvard University \\
  \texttt{finale@seas.harvard.edu}
  }

\begin{document}

\maketitle

\begin{abstract}
Global explanations of a reinforcement learning (RL) agent's expected behavior can make it safer to deploy. However, such explanations are often difficult to understand because of the complicated nature of many RL policies. Effective human explanations are often \textit{contrastive}, referencing a known contrast (policy) to reduce redundancy. At the same time, these explanations also require the additional effort of referencing that contrast when evaluating an explanation.  We conduct a user study to understand whether and when contrastive explanations might be preferable to complete explanations that do not require referencing a contrast.  We find that complete explanations are generally more effective when they are the same size or smaller than a contrastive explanation of the same policy, and no worse when they are larger.  This suggests that contrastive explanations are not sufficient to solve the problem of effectively explaining reinforcement learning policies, and require additional careful study for use in this context.
\end{abstract}

\section{Introduction}

Domain experts are increasingly using reinforcement learning (RL) agents to guide high-stakes decisions, in areas ranging from medical treatment to autonomous driving. Therefore, it is crucial to provide users with explanations of these agents' behavior. Global explanations of RL policies can be useful for answering big picture questions about the agent, like evaluating an agent's capabilities, choosing the most suitable agent for a given task, and deciding when to trust an agent's recommendation (\citep{huang_2018}, \citep{huang_2019}, \citep{amir_amir}, \citep{amir}, \citep{lage}). However, generating global explanations is challenging because of the complex computational techniques used by agents and the sheer size of the state space \citet{amir}. 

Contrastive explanations describe an event in reference to a known contrast that was expected to happen instead. These explanations are a natural and commonly-used form of human communication, and can potentially simplify descriptions by reducing redundancy \citet{miller}.  In this paper, we examine through a user study the use of \textit{contrastive} explanations as a strategy to reduce unnecessary complexity in global policy explanations. We find that complete explanations are generally preferred by at least one metric when they are the same size or smaller than the contrastive explanations, suggesting there is no innate preference for contrastive explanations in this context.  In the case where contrastive explanations are substantially smaller, there are no significant differences between the two explanation types. These results suggest that participants generally prefer complete explanations over contrastive ones, at least in cases when complete explanations are small enough to be understandable, and that further study of contrastive explanations is needed to understand how best to use them to interpret RL policies.

\begin{figure*}[h!]
\centering
\begin{subfigure}{.64\textwidth}
  \vspace{0pt}
  \centering
  \includegraphics[width=\linewidth]{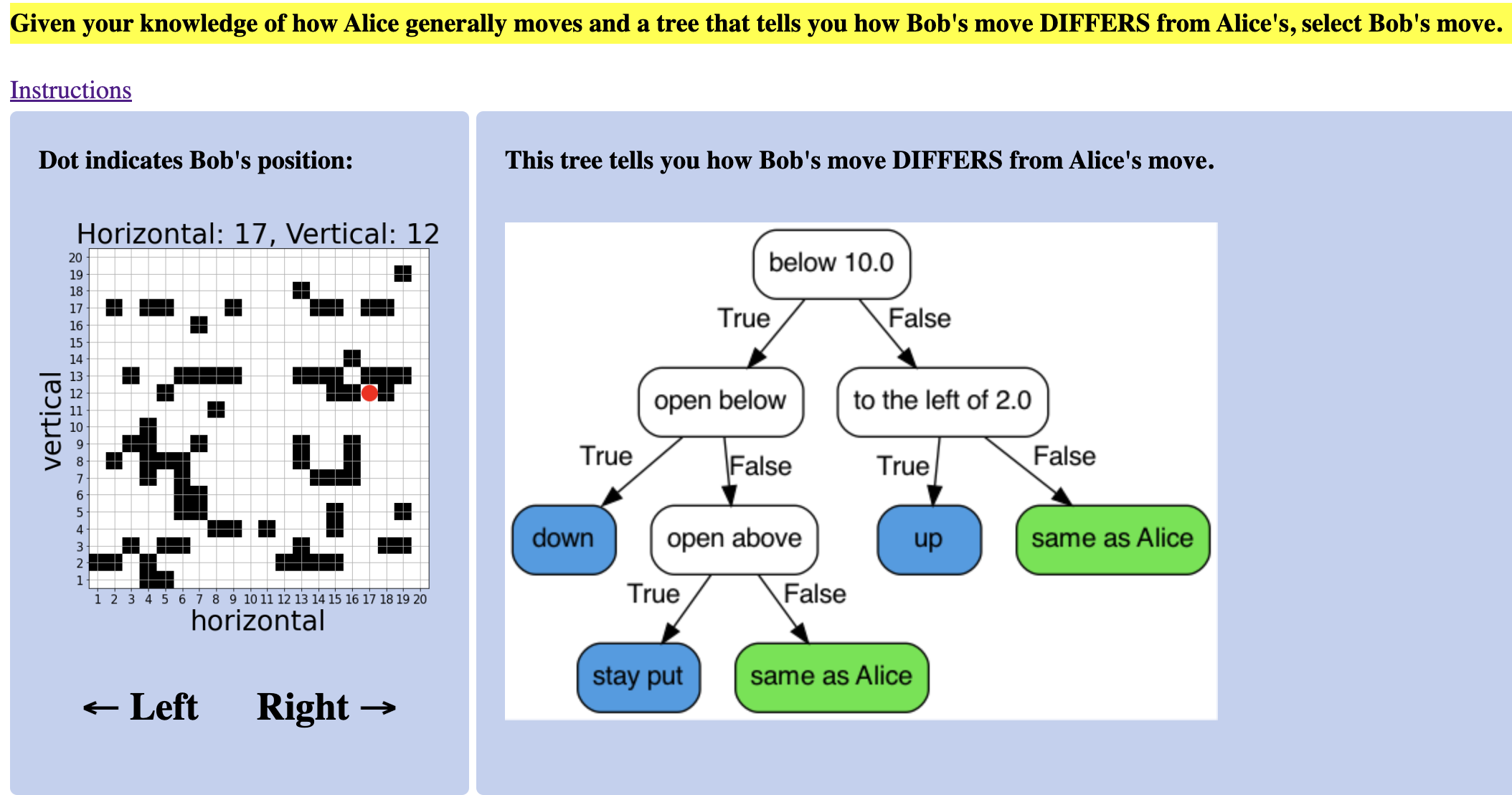}
  \caption{Our interface showing the task and a small, contrastive explanation.  The box on the left shows the maze domain where black squares are obstacles, and the red circle is the agent position, described in text above the map.  The explanation sometimes provides the action the agent takes in a state (blue leaves), and sometimes says the agent does the ``same as Alice'' (green leaves), which is the known contrast policy.}
  \label{fig:interface}
\end{subfigure}%
\begin{subfigure}{.35\textwidth}
  \vspace{0pt}
  \centering
  \includegraphics[width=\linewidth]{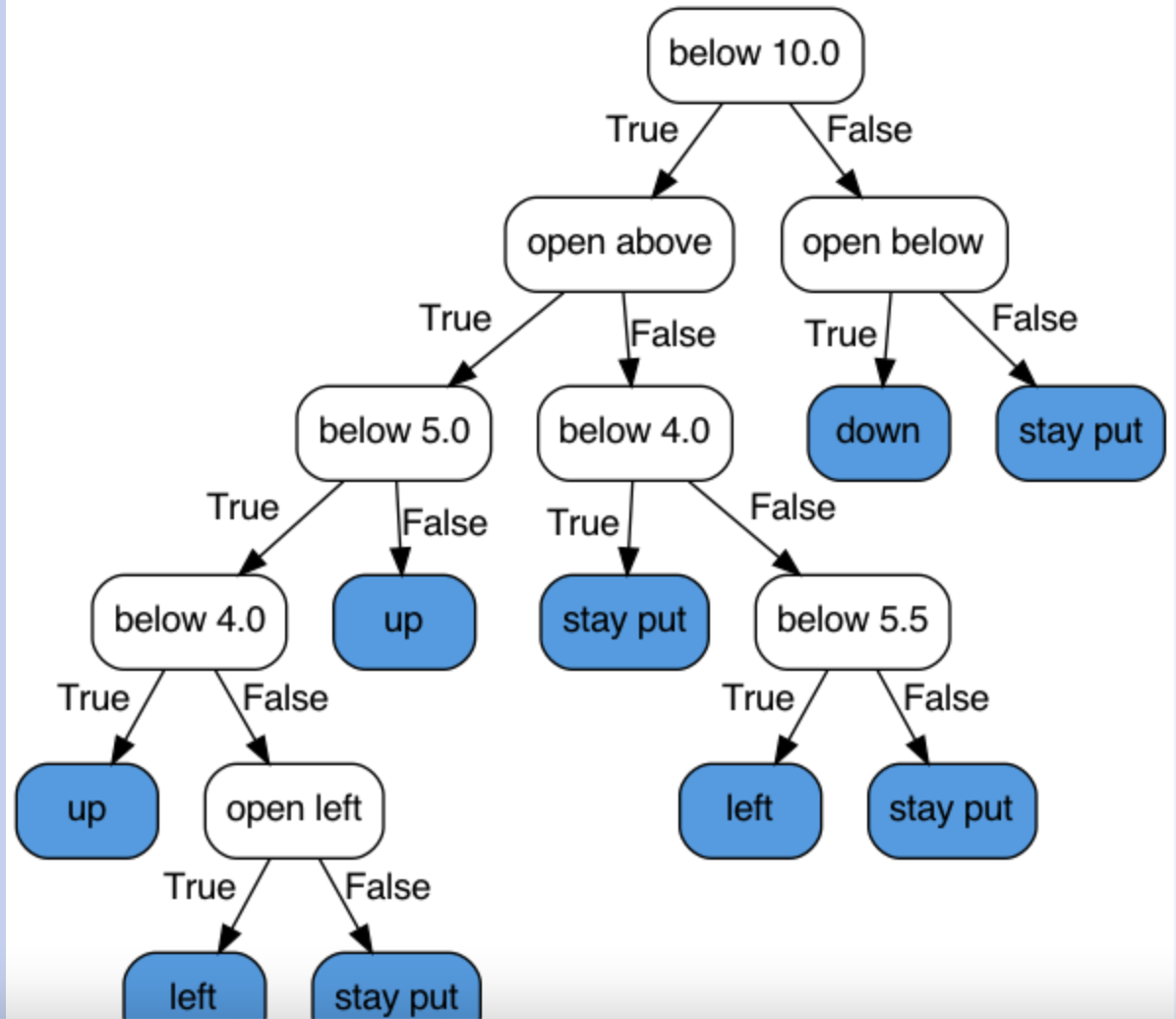}
  \caption{An example of a complete explanation which does not reference any contrast policy.  All leaf nodes are blue and evaluate directly to an action.  This particular example is a ``long'' explanation in contrast to the ``short'' explanation shown on the left.}
  \label{fig:long-complete-explanation}
\end{subfigure}
\end{figure*}

\section{Related Work} 

Cognitive science research suggests that when humans generate explanations, they often do so in the form of \textit{contrastive} explanations that describe why a particular event happened rather than a different, expected event--often called a contrast or a foil.  \citep{miller} relates this literature to explainable AI and highlights how contrastive explanations can often be more easily and succinctly produced than complete explanations, but how in order for them to be sensible, there must be significant similarities between the event that did happen and the foil used in explanation (see \citep{lipton_1990} for a more in depth discussion).  In this context, a complete explanation is an explanation that lists all of the causes of an event without making reference to a contrast.  Note that counterfactuals are related, but focus on alternative \textit{causes} of an outcome, rather than an alternative \textit{outcome} \citep{miller}.  We test whether and when contrastive explanations of reinforcement learning policies using a known behavior policy as a contrast are preferable to using a complete explanation of the RL policy.

Explanations of reinforcement learning policies are generally split into two categories: global explanations that explain the behavior of an agent in the entire state space at once, and local explanations that explain why an agent took a particular action in a particular state \citet{alharin}.  Global explanations of policies have been shown to be effective for tasks including choosing between multiple agents (\citep{amir_amir}, \citep{HUBER}), evaluating the importance of features towards agent decisions \citet{HUBER}, and anticipating agent actions (\citep{lage}, \citep{huang_2019}); and take various forms including decision trees (\citep{bastanirl}, \citep{roth}) and policy summaries that convey the agent's behavior as a set of state-action pairs (e.g. \citep{amir_amir}, \citep{huang_2019}).  Small decision trees can be readily understood by users (see e.g. \citep{freitas}), but can be difficult to train as effective RL policies requiring innovations like model distillation \citet{bastanirl} and Q-learning with constraints on tree complexity \citet{roth}.  In this work, we focus on global, decision tree-based explanations, however rather than exploring methods for training these, we use decision tree explanations as a test-bed to systematically compare contrastive and complete explanations.

Several types of contrastive and counterfactual explanations have been proposed for explaining RL policies.  Most previous work on these explanation types has either focused on comparing the consequences of taking one action over another (\citep{van_der_waa}, \citep{krarup}, \citep{lin}, \citep{madumal}) and learning local rather than global differences between policies (\citep{van_der_waa_trees}, \citep{hayes}), or on non-RL tasks altogether, such as classification (\citep{dhurandhar}).  Most are not evaluated through user studies, but \citep{van_der_waa} asks users to select their preferred explanation and \citep{madumal} presents users with a series of contrastive explanations of local agent actions, and then asks users to predict the agent's action at the next step and rate their trust of the agent.  While our task is similar, our explanations are global and convey the entire policy at once.  Finally, \citep{amir_2022} presents a global method for contrasting 2 learned RL policies and shows through user studies that the explanation contrasting these 2 policies allows users to more effectively pick the best policy than 2 complete summaries.  Our work differs from these in 2 key ways: 1) we evaluate the effectiveness of global contrastive explanations for anticipating an agent's behavior, which is important for human-AI collaboration, and 2) we use contrastive explanations to generate explanations of the policy using a contrast policy already familiar to the user rather than to compare 2 plausible RL agents.  

\section{Policy Explanation Methods}

We consider two types of global explanations of RL policies in this work: complete explanations that fully describe the agent's behavior in each state without reference to any other policy, and contrastive explanations that describe the agent's behavior in reference to a known contrast policy.  We define some background RL concepts below, then describe the key properties of each of these explanation types.  

\paragraph{Reinforcement Learning Background}

Our goal in this work is to explain policies of trained RL agents.  A policy $\pi(S) = A$ is a mapping between states $S$ and actions $A$ that the agent takes in those states.  An example of a state (as used this work) is the vertical and horizontal coordinates on a map, as well as whether there are obstacles in each of the four cardinal directions.  RL agents are generally trained so that the learned policy maximizes some specified reward function.  In this work, we consider deterministic policies, where the agent always takes the same action in a given state.

In our context, the contrastive summaries will also rely on a contrast policy, $\pi_{\texttt{ctrst}}$ that is not necessarily a trained RL agent, but is familiar to the user.  This is simply another mapping between states and actions that the user already knows.  One example of a $\pi_{\texttt{ctrst}}$ is the policy that the user would have followed in the absence of the RL agent.  

\paragraph{Complete explanations} 

Complete explanations fully describe the behavior of a trained RL agent (i.e. its policy).  They provide stand-alone descriptions of the agent's actions across the state space that can be used to understand how it will behave.  Formally, a complete policy explanation is a human-readable function 
\begin{equation}
\label{eqn:complete}
    e_{\texttt{cmplt}}(S) = \pi(S)
\end{equation}
that represents the RL agent's policy $\pi$.  We operationalize this function as a decision tree, which is relatively easy for users to understand and therefore commonly used in explainable RL \citet{alharin}.  See Figure~\ref{fig:long-complete-explanation} for an example of a complete explanation used in our experiment.  

\paragraph{Contrastive Explanations}

Rather than representing the trained RL agent's policy from scratch, contrastive explanations represent it in contrast to a known policy, $\pi_{\texttt{ctrst}}$. This is a natural explanation format for users, and an effective way to reduce redundancy when there is substantial overlap between what the user already knows and the agent's policy \citet{miller}.  Formally, a contrastive policy explanation is a a human-readable function
\begin{equation}
\label{eqn:contrastive}
e_{\texttt{ctrst}}(S, \pi_{\texttt{ctrst}}) = \begin{cases} 
    \pi(S) \texttt{ if } \pi(S) \neq \pi_{\texttt{ctrst}}(S) \\   
    \texttt{`Same as contrast' o.w.} 
\end{cases}
\end{equation}
that represents the RL agent's policy $\pi$.  When the contrast policy's action matches the agent's action in a given state, the explanation says to reference the contrast policy.  When the actions do not match, the explanation instead gives the action taken by the agent.  This explanation formalization fits into the framework of a contrastive explanation because the human-understandable function $e_{\texttt{ctrst}}$ is designed to answer only the question of why the agent took actions $\pi(S)$ rather than $\pi_{\texttt{ctrst}}(S)$ (in the cases where they differ).  As in the complete explanation case, we operationalize $e_{\texttt{ctrst}}$ as a decision tree.  See Figure~\ref{fig:interface}, right box, for an example of a contrastive explanation used in our experiment.

\section{Research Questions}

As each of these explanation methods has potential advantages and disadvantages, we aim to understand, through a user study, \emph{when} each of these methods should be employed.  Complete explanations are cognitively the most straightforward and stand-alone without needing to reference a contrast policy, while contrastive explanations align with how people are known to process explanations, and may be more concise in cases with substantial alignment between the agent's policy and the contrast policy.  These relative strengths and weaknesses lead us to our two main research questions:

\textbf{RQ1: When both explanation types are of the same size, does one or the other allow for better task performance?} This question allows us to determine whether one of the explanation types is more effective at facilitating task performance when all other factors are equal between them.  We measure task performance via the response time (the time that it takes the participant to perform the task).

\textbf{RQ2: When the explanation sizes differ, does the smaller explanation always lead to better task performance, or does the better performing explanation from RQ1 remain better?} This question is important because contrastive explanations are often more concise, as well as being a different form of summary.

\textbf{RQ3: Do the results hold across different metrics?}  We check to see if the patterns of perceived difficulty match the patterns in response time.

\section{Experimental Setup}

In this section, we describe the domain and method for generating the explanations and task inputs, then the experimental setup and analysis methods for the user study.

\paragraph{Domain}
\label{domain_subsection}

We created a simple maze domain where users could reason about the behavior of the agent and easily understand and learn the contrast policy we provided. The maze consists of a a 2-dimensional $20 \times 20$ grid with some regions blocked off as impassable, marked as black squares.  See Figure\ref{fig:interface}, the left box, for the visual representation of the domain given to users.  

The agent's state in this domain consists of its horizontal and vertical location on the grid.  Formally, the discrete states can be written as: $S = (x, y)$.  The agent's action space consists of the following 5 actions: move one unit up, down, right, or left, or stay put in its current position.  It cannot enter a blocked region or leave the grid.  

\paragraph{Contrast Policy} We defined one simple contrast policy for the experiment: ``If vertical position $y < 10$, move UP. Otherwise, move DOWN. If that move is blocked, STAY PUT."  We required users to memorize it at the start of the contrastive block of the experiment, and test their recall with an action prediction question halfway through the block.  $98.04\%$ of participants answered this question correctly, suggesting that they were able to recall the contrast.  

\subsection{Generating Explanations}
\label{policy_gen_subsection}

We followed a 2-stage process to generate both a complete and a contrastive explanation for the same policy, facilitating our analysis. To do this, we first randomly generated logical functions to assign actions across the state space--this was the policy. Then, then we trained decision trees on top of this function to accurately replicate the policy.  As these decision trees were trained to match the underlying policy in over $99\%$ of sampled states, we do not consider them to be approximations. However, in practice, one way to generate similar explanations is to use a model distillation approach on a black-box RL policy (see e.g. \citep{bastani}).

The final policies and explanations used in our experiment were chosen according to criteria defined in Section~\ref{sec:conditions}.  Below, we describe the approach for training the decision trees used as the explanations.  The sampling procedure used to generate candidate policies is described in Appendix~\ref{app:policies}.   

\paragraph{Generating Explanations from Policies}
\label{sec:generating-policies}

We derive the decision tree explanations from the policy representations described above by training a decision tree model to solve a supervised learning problem based on feature representations of each state, $\phi(S)$, and labels based on the action taken in $S$: $l_{\texttt{cmplt}}$ generated using Equation~\ref{eqn:complete} for the complete explanations, and $l_{\texttt{ctrst}}$ generated using Equation~\ref{eqn:contrastive} for the constrastive explanations.  

The features we used to represent the states consisted of the $x$ and $y$ coordinates of the state $S$, as well as whether the agent is blocked in each of the four cardinal directions.  The state representation $\phi(S)$ can be written as: $$s' = [x, y, b_\texttt{up}, b_\texttt{down}, b_\texttt{left}, b_{right}]$$ where $x, y \in [0, 20)$ and the $b_\texttt{direction}$ features are binary indicators that tell us whether there is an obstacle in the adjacent square in each of the 4 directions.  

To train the decision tree explanations, we sample N=$10,000$ states, $S = (x, y)$ uniformly at random and produce the label using the approach described above, with a 0.9/0.1 train/test split.  When training decision trees, we use the scikit-learn implementation \citet{scikit-learn} and set the following hyperparameters: the Gini impurity splitting criteria, unlimited max\_depth and max\_leaf\_nodes, and random\_state = 0.

We ensure that all decision tree summaries, complete and contrastive, have test accuracy of at least 0.99 by only considering those policies where this is true for both summary types.  This guarantees that our policy explanations are accurate representations of the policy, rather than approximations that may introduce additional challenges.  

\subsection{Task}

To measure participants' understanding, we use the task used in \citet{lage}--we ask them to predict the agent's behavior in a specified state.  We describe how these states were chosen in Appendix~\ref{app:states}  This measures to what extent they can anticipate how the agent will behave based on the provided explanation.

\subsection{Conditions}
\label{sec:conditions}

In order to answer our research questions about where each explanation type is effective, we test both types of explanations across 4 conditions that are determined by the sizes of both the complete explanation and the contrastive explanation for a given policy.  The 4 conditions are:
\begin{itemize}
  \item complete-small-contrast-small: both explanations are of the same size and small
  \item complete-small-contrast-large: the complete explanation is smaller than the contrastive explanation 
  \item complete-large-contrast-small: the contrastive explanation is smaller than the complete explanation 
  \item complete-large-contrast-large: both explanations are of the same size and large
\end{itemize}

We make the choice to test these 4 conditions that look at the cross product of both explanation type sizes for two reasons.  The first reason is that it allows us to reduce variance in the results and employ paired statistical tests for our main research questions, RQ1 and RQ2.  The second reason is that there may be systematic differences in the types of policies that tend to produce a long or a short explanation of each type, and we wish to avoid those interfering with our analysis.

We define explanation sizes based on both the maximum depth of the decision tree, and the number of leaves in the tree, whcih controls for path lengths and visual size of the trees.  We define a small summary as a tree of depth 3 (not including the root node) and either 4 or 5 leaves.  A large summary is a tree of depth 5 (not including the root node) and either 9 or 10 leaves.  See Figure~\ref{fig:interface} for an example of a small explanation, and Figure~\ref{fig:long-complete-explanation} for an example of a large explanation. 

\subsection{Selecting Policies and Explanations}

For each of the 4 conditions above, we generate 2 policies that satisfy the criteria and their corresponding summaries.  We additionally require that there exist states that satisfy the criteria described in Appendix~\ref{app:states}.  

\subsection{Metrics}

We record 3 metrics for each question to measure the how effective each of the explanations are based on the task.  These 3 metrics are: standardized  response time, subjective difficulty rating, and accuracy.  We standardize response time by subtracting off the participant specific mean response time.  Accuracy is measured as whether or not the action predicted by the user matches the true action taken by the agent at that state.  We asked participants how difficult it was to make each prediction (after the response time was stopped), on a likert scale from 1: `very easy' to 5 `very hard.'  In the experiment instructions, we told participants to focus on being accurate rather than fast, so we consider response time as the primary metric, and accuracy as a secondary metric.  

\subsection{Experimental Procedure}

We measured all independent variables within subjects, allowing us to reduce variance in our statistical analysis.  Participants were given 2 blocks of 4 questions, corresponding to the 2 explanation types.  Within those 4 questions were 1 from each of the 4 summary size conditions described in Section~\ref{sec:conditions}.  Policies were not repeated for a participant in order to avoid learning effects.

Participants were trained in the task before completing the study, and their understanding was evaluated with a set of practice questions.  Participants who failed to get these practice questions right in 2 tries were excluded from the analysis.  We also told participants that their primary goal was accuracy and their secondary goal was speed.  This results in relatively high task accuracies, motivating our choice to analyze response time.  See Appendix~\ref{app:accuracies} for details.  Additional details about the experimental procedure are given in Appendix~\ref{app:procedure}

\subsection{Recruiting Participants}

We recruited participants via Amazon Mechanical Turk.  We required participants have a HIT approval rate of 90 or greater and at least 500 HITs approved.  We paid participants \$5-\$7.  This study was approved by our institution's IRB.

We excluded participants who failed to get the practice questions right within 2 tries.  This criteria excluded $36/87$ participants, which is a substantial percent of respondents.  This means that these results may not generalize to the everyone in the general population, but should be representative of people who are more comfortable completing this task.  In a real-world setting, particularly a high-risk one, users are likely to have more training with the explanation system than we were able to provide in the context of this experiment.  We made a few additional exclusions based on highly abnormal response times.  We describe additional details of participant recruitment in Appendix~\ref{app:participants}.  

\subsubsection{Experimental Interface}

Figure~\ref{fig:interface} shows a screenshot of our interface with a small, contrastive explanation.  In the left box is the map of the domain with the specific state where the participant is asked to predict the agent's action marked with a red circle.  The coordinates are also listed in text at the top of the map.  The black squares are obstacles that cannot be moved through.  We annotated the map with the numbers and left and right arrows to facilitate reading the map.  

In Figure~\ref{fig:interface}, the right box shows the explanation of the policy with leaves that evaluate to an agent action directly marked in blue, and leaves that evaluate to the contrast policy marked in green.  True and False marked on the decision tree arrows facilitate navigating the tree.  Figure~\ref{fig:long-complete-explanation} shows the explanation for a long complete summary.  All leaves evaluate directly to an action, so are marked in blue.  Otherwise, the explanation is presented identically in both explanation types.

Below the interface shown in Figure~\ref{fig:interface}, we ask participants to predict the action in the red state in multiple choice format from the 5 action possibilities (up, right, down, left, stay put).  See Appendix~\ref{app:interface} for additional details.

\subsubsection{Analysis}

We ran statistical tests on our collected data to answer our research questions.  We ran paired-sample 2-sided t-tests to compare the standardized response times for the 2 summary types across the 4 conditions as standardized response time is a continuous measure.  To compare perceived difficulty, we chose the Wilcoxon signed-rank test which accounts for paired samples.  All tests were implemented using the scipy stats software.

We set the threshold for statistical significance at $p=0.05$.  In order to compare for multiple hypothesis testing, we used a Benjamini-Hochberg correction for multiple hypothesis testing for 16 tests we ran.  Significant results are starred in figures and described in the text.  Appendix~\ref{app:stats} includes details about the statistical results including p-values.  

\section{Results}

We describe the results of our research questions \textbf{RQ1}, \textbf{RQ2} and \textbf{RQ3} below.   We find that that complete explanations allow for quicker task performance than contrastive explanations when both are large or when the complete explanations are smaller.  Participants also generally perceived complete explanations as less difficult when they were small.  Contrastive explanations never significantly outperformed complete explanations, even when they were substantially smaller.

Figure\ref{fig:main_rt_result} shows the difference in standardized response times for the cases with large and small complete and contrastive explanations.
 
\begin{table}
    \begin{minipage}{0.4\linewidth}
        \centering
        \includegraphics[width=\linewidth]{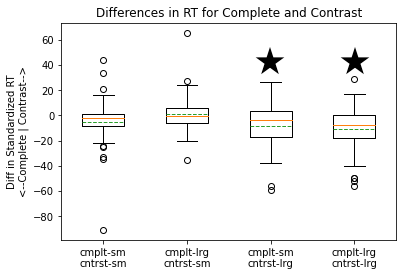}
        \captionof{figure}{This figure shows the distribution of differences in standardized response time (RT) between the 2 explanation types for each participant in the 4 explanation size conditions. The box shows the 1st-3rd quartile with the median marked in orange and the mean in green (dotted). Negative differences mean lower standardized RT for complete explanations, and positive for contrastive explanations.  Complete explanations have significantly lower (marked with a star) standardized RT in both conditions where the contrastive explanation is large.}
        \label{fig:main_rt_result}
    \end{minipage}
    \begin{minipage}{0.6\linewidth}
        \centering\includegraphics[width=1.\linewidth]{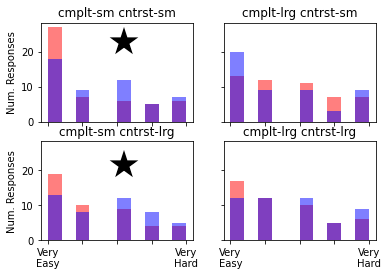}
        \caption{This figure shows the distribution of perceived difficulty scores for complete (red) and contrastive (blue) explanations in the 4 explanation size conditions.  Purple indicates overlap between the two explanation types.  For both cases where the complete explanation is small, it is perceived as significantly (star) less difficult than the contrastive explanations, while in the other case, the differences are not significant.}
        \label{fig:main_difficulty_result}
    \end{minipage}\hfill
\end{table}

\textbf{RQ1 Conclusion: When both explanation types are large, complete summaries have significantly lower standardized response time, while when the explanation types are both small, the trend is consistent but not statistically significant.}  Figure\ref{fig:main_rt_result} shows the difference in standardized response times for the complete and contrastive explanations in each of the 4 explanation-size conditions.  In Figure\ref{fig:main_rt_result}, the complete-large-contrast-large condition has a difference in standardized response time significantly below zero ($p=7.3757e^{-3}$, $t=-4.3289$), suggesting that, when both summary types are long, a complete explanations allows participants to perform the task more quickly.  In the complete-small-contrast-small condition, we see a similar trend with the difference in standardized response times below zero suggesting that a small complete explanation is preferred to a small contrastive explanation, however the result is not statistically significant ($p=5.2015e^{-2}$, $t=-1.9905$).  Overall, these results suggest that, when explanation sizes are similar, a complete explanation allows for more efficient task completion than a contrastive explanation, and the difference is more exaggerated as explanation sizes grow.  
 
\textbf{RQ2 Conclusion: When the complete explanation is smaller than the contrastive explanation, it has significantly lower standardized response time, but the reverse is not true, suggesting there is no substantial difference in response time between a small contrastive explanation and a large complete explanation.}  In Figure\ref{fig:main_rt_result}, in the complete-small-contrast-large condition, the difference in standardized response times skews lower than 0, indicating that response times are lower for the smaller complete explanations than the large contrastive ones.  This result is statistically significant ($p=4.0023e^{-3}$, $t=-3.0337$).  In the reverse case, the complete-large-contrast-small condition, the difference in standardized response time is not significant ($p=6.8238e^{-1}$, $t=0.411$), and visually, it appears to be closely centered on zero.  This suggests that a small contrastive explanation and a large complete explanation take similar amounts of time to process.

\textbf{RQ3 Conclusion: Perceived difficulty results suggest a lower perceived difficulty for complete explanations when they are small, regardless of contrastive explanation size.} Figure~\ref{fig:main_difficulty_result} shows histograms of subjective difficulty ratings for both explanation types in each of the 4 explanation-size conditions.  In the complete-small-contrast-small and complete-small-contrast-large conditions, there is a statistically-significantly lower perceived difficulty for the complete summary ($p=1.8259e-03$, $W=15.0$; $p=1.1171e-02$, $W=64.5$).  The differences in perceived difficulty are not significant for the 2 other conditions.  When both explanation sizes are large, the trend is towards complete explanations having a lower perceived difficulty, which is consistent with the other findings.  When the complete explanation is large and the contrastive explanation is small, the results are mixed.

\section{Discussion}

Our results suggest that complete explanations may generally be preferrable to contrastive explanations as they are significantly better in either response time or perceieved difficulty for all conditions where the complete explanation is the same size or smaller than the contrastive explanation.  While these metrics do not perfectly align, they capture slightly different aspects of the problem, both of which are important.

When the contrastive explanation is shorter than the complete explanation, they appear to perform quite similarly across metrics.  This raises the question of whether, with a complete summary that is much larger, we might see comparatively better task performance for the contrastive explanation.  In this experiment, we constrained the explanations to sizes that could be reasonably displayed on the screen, and that were not too challenging for participants to answer correctly.  It seems likely that larger complete summaries will be really difficult for participants to understand, shifting the balance towards the contrastive summaries.   

While it is not clear from our results exactly why the contrastive explanations generally performed worse, particularly given that they are commonly used in human interactions, it may be due to the additional cognitive load of keeping the contrast policy in mind, or to other unexplored factors.  Whether there are ways to present contrastive explanations to increase their effectiveness remains an open question.  

\section{Conclusion}

Contrastive explanations are a natural form of explanation in human communication, and can result in less complex explanations in cases where the contrast shares many of the features of the event to be explained.  While this suggests they may be useful for providing more effective explanations of reinforcement learning policies, the results of our user study suggest that complete explanations are often preferred and never worse, at least in cases where they are reasonably sized.  Further work exploring the use of contrastive explanations for reinforcement learning policies should be careful to identify and mitigate the factors that cause them to be more challenging to work with than complete explanations.

\begin{ack}
This material is based upon work supported by the National Science Foundation under Grant No. IIS-2107391.  Any opinions, findings, and conclusions or recommendations expressed in this material are those of the author(s) and do not necessarily reflect the views of the National Science Foundation.  IL was funded by NSF GRFP (grant no. DGE1745303).
\end{ack}

\bibliographystyle{plainnat}
\bibliography{sample}

\appendix

\section{Sampling Policies}
\label{app:policies}

To generate candidate policies, we specified a functional form with parameters that could be randomly sampled to generate policies with distinct properties. Each policy consisted of at most four conditional statements, each corresponding to an action.  

Each conditional statement corresponds to an `and' or an `or' of 2 thresholds $t_1$ and $t_2$. Each threshold can be either an upper bound or a lower bound, each threshold can apply to either the $x$ or $y$ coordinates of the map, and each threshold evaluates to True or False for a given state $S = (x, y)$. A condition could be written as $x < t_1 \vee x > t_2$, or $x > t_1 \wedge y < t_2$, just to give two examples.

Using this procedure to generate conditional statements given 2 sampled thresholds, $t_1$ and $t_2$, we can generate multiple conditional statements and use them to assign actions, also randomly sampled, to different parts of the state space.  All policies are sampled according to the form described in Algorithm~\ref{alg:policy-form} where the sampled parameters are: $\{t^1_1, t^1_2, t^2_1, t^2_2, t^3_1, t^3_2, a_1, a_2, a_3, a_4\}$.



While other policies can be specific in this domain, we find that this space of policies is sufficiently expressive to generate policies that meet the criteria defined for each of our conditions.  

\begin{algorithm}
\begin{algorithmic}
\If{$cond_1(s, t_1^1, t_2^1)$}
    \If{$cond_2(s, t_1^2, t_2^2)$}
        \State \Return $a_1$
    \Else{}
        \State \Return $a_2$
    \EndIf
\ElsIf{$cond_3(s, t_1^3, t_2^3)$}
    \State \Return $a_3$
\Else{}
    \State \Return $a_4$
\EndIf
\end{algorithmic}
\caption{Form of the logical rules used to generate policies used in our experiments.  Parameters $\{t^1_1, t^1_2, t^2_1, t^2_2, t^3_1, t^3_2, a_1, a_2, a_3, a_4\}$ are all randomly sampled to generate a set of random policies to select questions from.}
\label{alg:policy-form}
\end{algorithm}

\paragraph{Selecting States for the Task}
\label{app:states}

The task requires specifying a state in which to predict the action.  We set some requirements on these states to guarantee that the answer is well specified, the questions are consistent with the conditions, and to reduce the impact of learning effects.  To guarantee the question is well specified, we require that the state is not blocked, that it is not directly on the border of the grid (where the policy may be ambiguous to users), and the state is at least one away from the $x=10$ line as this is where the contrast policy action changes sign and we did not want participants to be required to remember exactly where this occurred (i.e. is it $> 10$ or $\geq 10$?).  To guarantee that the questions are consistent within a condition, we require the decision path of the state within the explanation tree to be within 1 of the maximum depth (so we don't have, for e.g., trees of size 5 and states with decision paths of length 2).  To guard against participants learning a prior on the correct response for the contrastive explanation, we require one of the two questions in each condition to evaluate to the contrast policy with the contrastive summary, and the other to evaluate directly to an action.  We used the same state for both explanation types for a given policy.

\section{Experimental Procedure}
\label{app:procedure}

In order to train participants in the task, we gave participants a general set of instructions at the beginning of the task that included a description of the domain, an explanation of how to read decision trees, and instructions for how to complete the task.  The instructions for each specific explanation type, and the contrast policy for the contrastive questions were shown directly before the start of their respective question blocks.  At the outset of the task, we told participants that their primary goal was accuracy and their secondary goal was speed.
Participants were then given a sample policy with 3 practice questions (each a state where they must prediction the action), and were required to get all 3 questions correct before moving on.  If participants did not correctly answer all 3 questions on the first try, they were given a second try with a new policy.  Participants who required more than 2 tries to get either of the practice questions right were excluded from the analysis.  In addition, participants were given a set of 5 practice questions about the contrast policy after reading its description that they were required to answer correctly before moving on.  This was repeated, along with the instructions, until they did so.  This question was not used to exclude participants.

When completing the task, participants were first asked to make the prediction.  Afterwards, they were given a pop-up where they were asked about the difficulty of the question.  After that, they received a second pop-up telling them whether they answered the question correctly or not.  Finally, some participants were asked to describe the policy in words based on the explanation, although we removed this question after the first round of data collection as this the question felt ill-specified, even to participants who otherwise performed well on the task.

In addition to the task, we asked participants several questions about their demographics, and their experience with the task.  At the start of the study, we asked a series of demographic questions.  A single memory-check question was asked in the middle of the contrast block about the contrast policy to verify that participants remembered the contrast policy.  If they answered incorrectly, they were shown the contrast policy again and given another try.  After each block of questions, we asked participants about their experience with the explanation type, and at the end we asked them to compare the 2 types.  Finally, at end, we ask a free-text question about participants' experience with the survey.

\section{Recruiting Participants}
\label{app:participants}

We recruited participants via Amazon Mechanical Turk.  For an original set of 27 responses that included a free-text description of the explanation, we required participants have a HIT approval rate of $> 95$ and at least 1000 HITs approved.  For the remaining respondents where the free-text questions were not included, we lowered the HIT approval rate to $> 90$ and the number of HITs approved to $> 500$ as we determined that the difficulties were coming from the ill-specified nature of the free text question rather than participant qualifications.  We paid participants who completed the version with free-text descriptions of the summaries \$7 dollars, and the participants who completed the version without free-text summaries \$5.  This study was approved by our institution's IRB.

We excluded participants based on the practice question criteria described above (requiring more than 2 tries to get either of the sets of practice questions right, excluding those about the contrast policy).  This criteria excluded $36/87$ participants, which is a substantial percent of respondents.  This means that these results may not generalize to the everyone in the general population, but should be representative of people who are more comfortable completing this task.  In a real-world setting, particularly a high-risk one, users are likely to have more training with the explanation system than we were able to provide in the context of this experiment.

We additionally exclude a small number of responses based on response time as there was a long tail of responses taking minutes to make the predictions.  We set this threshold at 2 minutes to exclude responses where the participant likely got distracted while answering the question.  We additionally excluded the paired response (i.e. the response in the same condition with the other explanation type) to facilitate the statistical analysis.  We excluded 8 pairs of questions out of 204 pairs based on this criteria.  We note that setting this threshold higher or removing it does not change the statistical results.  We did not do any exclusions based on accuracy in the main task, but note that accuracies are generally high in the experiment. 

\section{Interface}
\label{app:interface}

\begin{figure}[]
  \vspace{0pt}
  \centering
  \includegraphics[width=\linewidth]{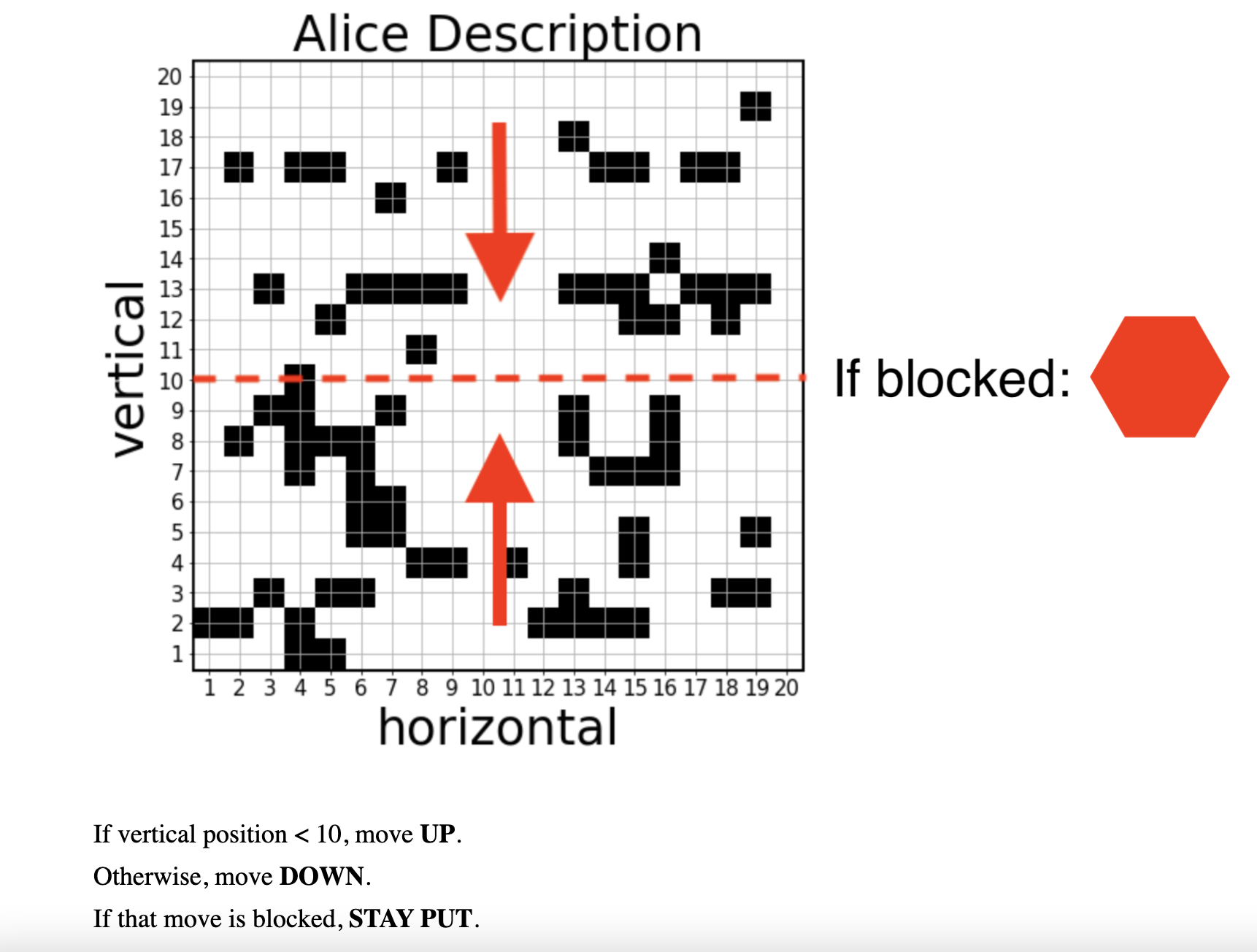}
  \caption{The representation of the contrast policy shown in our experiment.  This includes both a textual description and a visual summary of the policy.  In the instructions, the contrast policy is described as a policy belonging to a player in a maze game named Alice.}
\label{fig:alice-policy}
\end{figure}

\begin{figure}[]
  \vspace{0pt}
  \centering
  \includegraphics[width=\linewidth]{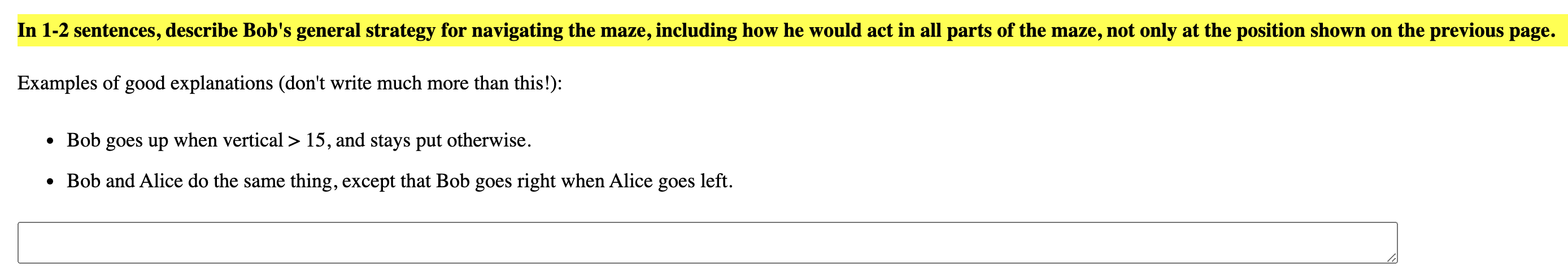}
  \caption{This is the interface used in our experiment asking the policy description question asked in a small subset of our responses.  We additionally show the explanation, and the maze map without marking any state in red.  For context, we suggest a size for the description, and give some examples.}
  \label{fig:global-question}
\end{figure}

We show additional screenshots of the interface showing the explanation of the contrast policy in Figure~\ref{fig:alice-policy}, and the policy description question asked in a subset of the initial surveys in Figure~\ref{fig:global-question}. 

\section{Experiment Accuracies}
\label{app:accuracies}

Accuracies are generally high across conditions with no significant differences.  This motivates our choice to analyze response time rather than accuracy.

\begin{table}[h!]
\centering
\setlength\tabcolsep{4pt}
\begin{tabular}{lrrrr}
Explanation & cmplt-sm & cmplt-lrg & cmplt-sm & cmplt-lrg \\
 type    & cntrst-sm & cntrst-sm & cntrst-lrg & cntrst-lrg\\\hline
Complete  & 0.8627 & 0.84 & 0.8261 & 0.9\\
Contrastive  & 0.8039 & 0.92 & 0.7174 & 0.8\\
\end{tabular}
\caption{Accuracies for each summary type for each condition.  There are no statistically significant differences between conditions.}
\label{tab:accuracies}
\end{table}

\section{Statistical Tests}
\label{app:tests}

We report significant tests and corresponding statistics in the main body of the paper, but we include all of the test outcomes in Table~\ref{tab:statistical-results} here for additional information about multiple hypothesis testing-corrected significance thresholds, and statistics and p values for tests that were not significant.  Note that several of these tests were run, but are not included in the results of this version, however we still consider them in the Bonferonni correction.  

\begin{table*}[ht!]
\begin{tabular}{lrrrl}
Test & P Value & Threshold & Statistic & Test \\
Diff. in standardized rt: cmplt-sm-cntrst-sm & 5.2015e-02 & 2.5e-02 & -1.9905 & 2-sided t test\\
Diff. in standardized rt: cmplt-lrg-cntrst-sm & 6.8283e-01 & 5.e-02 & 0.411 & 2-sided t test\\
Diff. in standardized rt: cmplt-sm-cntrst-lrg & \textbf{4.0023e-03} & 1.5625e-02 & -3.0337 & 2-sided t test\\
Diff. in standardized rt: cmplt-lrg-cntrst-lrg & \textbf{7.3757e-05} & 9.3750e-03 & -4.3289 & 2-sided t test\\
Diff. in accuracy: cmplt-sm-cntrst-sm & 5.4883e-01 & 4.3750e-02 & 4.0 & mcnemar\\
Diff. in accuracy: cmplt-lrg-cntrst-sm & 2.8906e-01 & 4.0625e-02 & 2.0 & mcnemar\\
Diff. in accuracy: cmplt-sm-cntrst-lrg & 1.7969e-01 & 3.125e-02 & 2.0 & mcnemar\\
Diff. in accuracy: cmplt-lrg-cntrst-lrg & 2.2656e-01 & 3.4375e-02 & 3.0 & mcnemar\\
Diff. in subjective difficulty: cmplt-sm-cntrst-sm & \textbf{1.8259e-03} & 1.25e-02 & 15.0 & wilcoxon\\
Diff. in subjective difficulty: cmplt-lrg-cntrst-sm & 2.3269e-01 & 3.7500e-02 & 90.5 & wilcoxon\\
Diff. in subjective difficulty: cmplt-sm-cntrst-lrg & \textbf{1.1171e-02} & 1.8750e-02 & 64.5 & wilcoxon\\
Diff. in subjective difficulty: cmplt-lrg-cntrst-lrg & 1.097e-01 & 2.8125e-02 & 78.5 & wilcoxon\\
Complete size impact & 4.8119e-02 & 2.1875e-02 & -1.9888 & independent t-test\\
Contrastive size impact & \textbf{3.9552e-06} & 3.125e-03 & -4.7489 & independent t-test\\
Contrast reference impact & 6.0267e-01 & 4.6875e-02 & 0.5214 & independent t-test\\
Explanation preference & \textbf{1.4192e-05} & 6.25e-03 & 18.8431 & chi square\\
\end{tabular}
\caption{This table shows the results of the statistical tests ran for each analysis including the type of test, the test statistic, the p value, and the threshold for significance based on the Benjamini-Hochberg correction for multiple hypothesis testing.  There are 6 significant results, generally suggesting that complete explanations are less cognitively demanding than contrastive explanations.}
\label{tab:statistical-results}
\end{table*}

\end{document}